\documentclass[10pt,twocolumn,letterpaper]{article}

\usepackage{3dv}
\usepackage{times}
\usepackage{epsfig}
\usepackage{graphicx}
\usepackage{tikz}
\usepackage{comment}
\usepackage{amsmath,amssymb} % define this before the line numbering.
\usepackage{color}
\usepackage{bbm}
\usepackage{multirow}
\definecolor{mygray}{gray}{.9}

\usepackage{colortbl}
\usepackage{blindtext}
\threedvfinalcopy % *** Uncomment this line for the final submission

\def\threedvPaperID{****} % *** Enter the 3DV Paper ID here

\ifthreedvfinal\fi
\begin{document}

%%%%%%%%% TITLE
\title{Bidirectional Feature Globalization \\ for Few-shot Semantic Segmentation of 3D Point Cloud Scenes}

\author{Yongqiang Mao$^{1,\dagger}$, Zonghao Guo$^{1,\dagger}$, Xiaonan Lu$^{1}$, Zhiqiang Yuan$^{1}$, Haowen Guo$^{2,}$\thanks{Corresponding author. $\dagger$Equal contribution.}\\
$^{1}$University of Chinese Academy of Sciences, Beijing China\\
$^{2}$Wuhan University, Wuhan China\\
{\tt\small mao.wingkeung@gmail.com, guozonghao19@mails.ucas.ac.cn, ghw@whu.edu.cn}
% For a paper whose authors are all at the same institution,
% omit the following lines up until the closing ``}''.
% Additional authors and addresses can be added with ``\and'',
% just like the second author.
% To save space, use either the email address or home page, not both
% \and
% Second Author\\
% Institution2\\
% First line of institution2 address\\
% {\tt\small secondauthor@i2.org}
}

\maketitle
\thispagestyle{empty}

\begin{abstract}
Few-shot segmentation of point cloud remains a challenging task, as there is no effective way to convert local point cloud information to global representation, which hinders the generalization ability of point features.
In this study, we propose a bidirectional feature globalization (BFG) approach, which leverages the similarity measurement between point features and prototype vectors to embed global perception to local point features in a bidirectional fashion.
With point-to-prototype globalization (Po2PrG), BFG aggregates local point features to prototypes according to similarity weights from dense point features to sparse prototypes.
With prototype-to-point globalization (Pr2PoG), the global perception is embedded to local point features based on similarity weights from sparse prototypes to dense point features. 
The sparse prototypes of each class embedded with global perception are summarized to a single prototype for few-shot 3D segmentation based on the metric learning framework.
Extensive experiments on S3DIS and ScanNet demonstrate that BFG significantly outperforms the state-of-the-art methods.
\end{abstract}

\section{Introduction}
\label{sec:intro}
\begin{figure}[t]
    \begin{center}
    \includegraphics[width=0.95\linewidth]{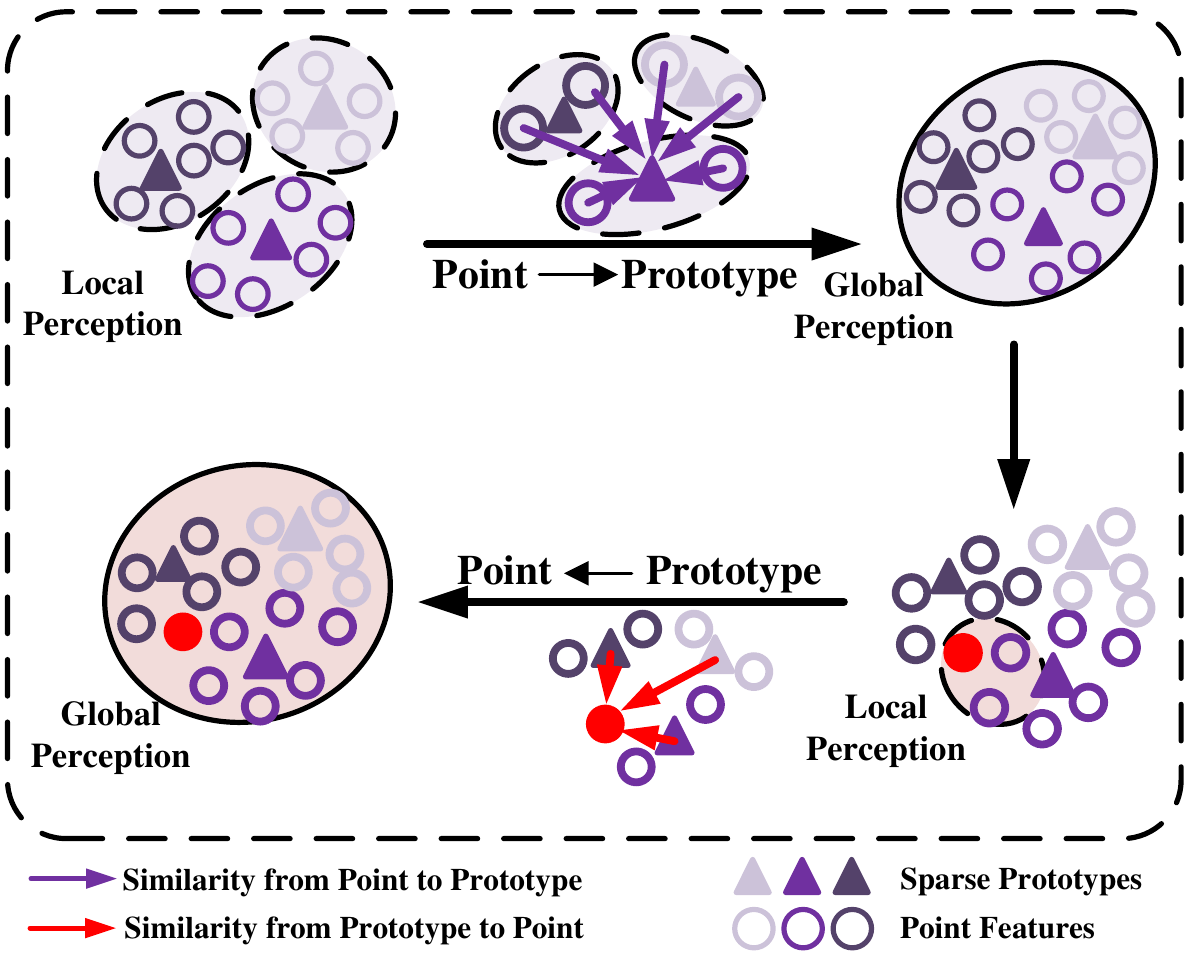}
    \end{center}
    \setlength{\abovecaptionskip}{0pt}
    \caption{Overview of our bidirectional feature globalization (BFG). 
    BFG first realizes the prototype feature globalization by integrating dense point features through similarity weights (Upper). 
    Point feature globalization is then performed by integrating sparse prototypes through similarity weights (Lower). Equipped with BFG, both prototypes and point features are endowed with global perception, which facilitates few-shot 3D segmentation.}
\label{motivation}
\end{figure}

Thanks to the powerful representation capabilities of Convolutional Neural Networks and the open source of numerous point cloud annotation datasets, we have witnessed unprecedented progress~\cite{qi2017pointnet,boulch2020convpoint,jiang2018pointsift,wang2019dynamic,ye20183d,mao2022beyond,mao2022semantic} in point cloud segmentation. However, annotating large-scale point cloud datasets is laborious and extensive, which hinders the application of point cloud segmentation in various scenarios.

In recent years, few-shot 3D point cloud segmentation~\cite{zhao2021few} is explored. Given base classes with sufficient training data and new classes of few supervisions, this task aims to generalize the 3D representation model initialized upon the base classes to new classes. Early researches simply imitate the few-shot segmentation methods from 2D to 3D tasks. For example, the single prototype method~\cite{dong2018few} leverages global average pooling to produce 3D semantic prototypes, which are used to classify point sets in a metric learning framework. The multi-prototype method~\cite{zhao2021few} generates prototypes by aggregating point features from different object parts to improve the semantic representation.

In spite of the substantial progress, existing methods are impacted by serious false segmentation when the instance consists of complex parts. By our investigation, we realize that the false segmentation is caused by the locality of point convolution, which lacks the ability to capture global feature perception. Such global perception is crucial to produce correct segmentation when deformation or scale variation occur.

In this paper, we focus on designing a bidirectional feature globalization (BFG) approach (Fig.~\ref{motivation}) to regularize the training procedure of semantic prototypes and endow each point feature and prototype the global feature perception. BFG defines a bidirectional module which uses the dense local point features to generate sparse global prototypes and then leverages the global prototypes to guide globalization of local point features. With such a bidirectional module, both point features and prototypes are endowed with global perception.

Specifically, the proposed approach is rooted in a metric learning framework, which consists of two branches (support branch and query branch), Fig.~\ref{framework}. 
The support branch is responsible for the globalization of prototypes and support point features, which goes through two modules in order: Point-to-Prototype Globalization (Po2PrG) and Prototype-to-Point Globalization (Pr2PoG).
Given the sparse prototypes initialized by the sparse prototype generation (SPGen) module, Po2PrG and Pr2PoG perform globalization on prototypes and point features in a bidirectional fashion according to the similarity weights between sparse prototypes and dense point features.
Finally, sparse prototype assembly (SPA) is carried out to obtain the optimal prototype representation of support point features for metric learning. 
The query branch generates the similarity maps between the prototypes from the support branch and the query point features to obtain point cloud segmentation results.

To conclude, the main contributions of our BFG are summarized as follows:
\begin{itemize}
    \item We propose bidirectional feature globalization (BFG), defining a simple-yet-effective way to embed global perception to local point features and their prototypes in a mutual enhancement fashion.
    
    \item We design point-to-prototype globalization (Po2PrG) and prototype-to-point globalization (Pr2PoG) modules based on the similarity weights, which activate the global perception of prototypes and point features, respectively. 
    
    \item By assembling sparse prototypes embedded with global perception, we achieve a new state-of-the-art performance on the popular S3DIS and ScanNet datasets.
\end{itemize}

%------------------------------------------------------------------------
\section{Related Work}

%------------------------------------------------------------------------
\textbf{Point Cloud Segmentation.}
Point Cloud semantic segmentation for point-wise classification of a class of instances has been extensively studied.
PointNet~\cite{qi2017pointnet} learns the features of each point independently through MLP, and utilizes a symmetric function (such as max pooling) to solve the disorder problem while aggregating global features.
On this basis, a large number of point-based methods~\cite{qi2017pointnet++,liu2019densepoint,wang2019dynamic} have sprung up. 
Point-wise MLP methods ~\cite{qi2017pointnet++,jiang2018pointsift,zhao2019pointweb,hu2020randla} employ shared MLP as the basic block of the network for feature extraction. 
Point Convolution methods ~\cite{liu2019densepoint,atzmon2018point,boulch2020convpoint,wu2019pointconv,li2018pointcnn,xu2021paconv,thomas2019kpconv,liu2019dynamic} aim to extract high-quality point features and learn local relationships by designing efficient point convolution operators. 
Graph-based methods~\cite{wang2019dynamic,wang2019graph,landrieu2018large} aim to learn the spatial geometric features of points through constructing graphs inside point sets and designing novel graph convolutions.

However, these methods heavily depend on large-scale training sets and are incapable of generalizing to new classes, which limits the application of many real-world scenarios. 
To improve the generalization capability of 3D representation models, few-shot segmentation task of point cloud has been the research focus of the community. 

%------------------------------------------------------------------------
\textbf{Few-shot Learning.}
Current methods of few-shot learning mainly concentrate on metric learning~\cite{vinyals2016matching,sung2018learning,zhang2020sg,zhang2019canet,shaban2017one} and meta-learning~\cite{wang2016learning,ravi2016optimization,finn2017model,jamal2019task}. 
The methods based on metric learning mainly focus on employing distance metric to predict whether two regions belong to the same class.
The main idea of meta-learning based methods is to specify an optimization procedure or loss function to gain the ability to learn faster and adapt to new classes. 
The effectiveness of the prototype concept for few-shot learning is demonstrated~\cite{vinyals2016matching,zhang2019canet} in various metric learning frameworks.
Inspired by these, the methods of prototype learning is widely adopted in the few-shot segmentation task, which greatly reduces the computational budget while maintaining high performance.

%------------------------------------------------------------------------
\textbf{Few-shot 3D Point Cloud Segmentation.}
Current few-shot segmentation methods~\cite{dong2018few,zhao2021few} of point cloud largely follow the metric learning framework, which learns semantics from the support point sets. Such stores the semantics in the form of prototype vectors, which are generalized to segment the query point sets. 
ProtoNet~\cite{dong2018few} uses a single prototype to centrally express the features of each semantic class in the support point sets. It designs a mask average pooling strategy to generate prototype vectors, and then applies a similarity measurement function to build the relationship between the prototypes and the features of query point sets. 
MPTI~\cite{zhao2021few} introduces a method of transductive learning to predict the semantic classes of the query point sets based on the prototypes. It also extracts multiple prototypes of the support point features to better represent the rich foreground semantic.

In spite of the substantial progress, existing methods are impacted by serious false segmentation when there exists deformation and/or scale variation. By our investigation, we realize that the false segmentation is caused by the locality of point convolution, which lacks the ability to capture global feature perception.
The multi-prototype method~\cite{zhao2021few} took a step to alleviate this. However, it requires the significant increase of the number of prototypes, which aggregates the computational complexity. In this paper, we propose the bidirectional feature globalization (BFG) approach, which aims to globalize features to obtain the optimal representation with a single prototype vector of each class.

%------------------------------------------------------------------------
\section{Method}

%-------------------------------------------------------------------------
\subsection{Overview}
\begin{figure*}
    \begin{center}
    \includegraphics[scale=.26]{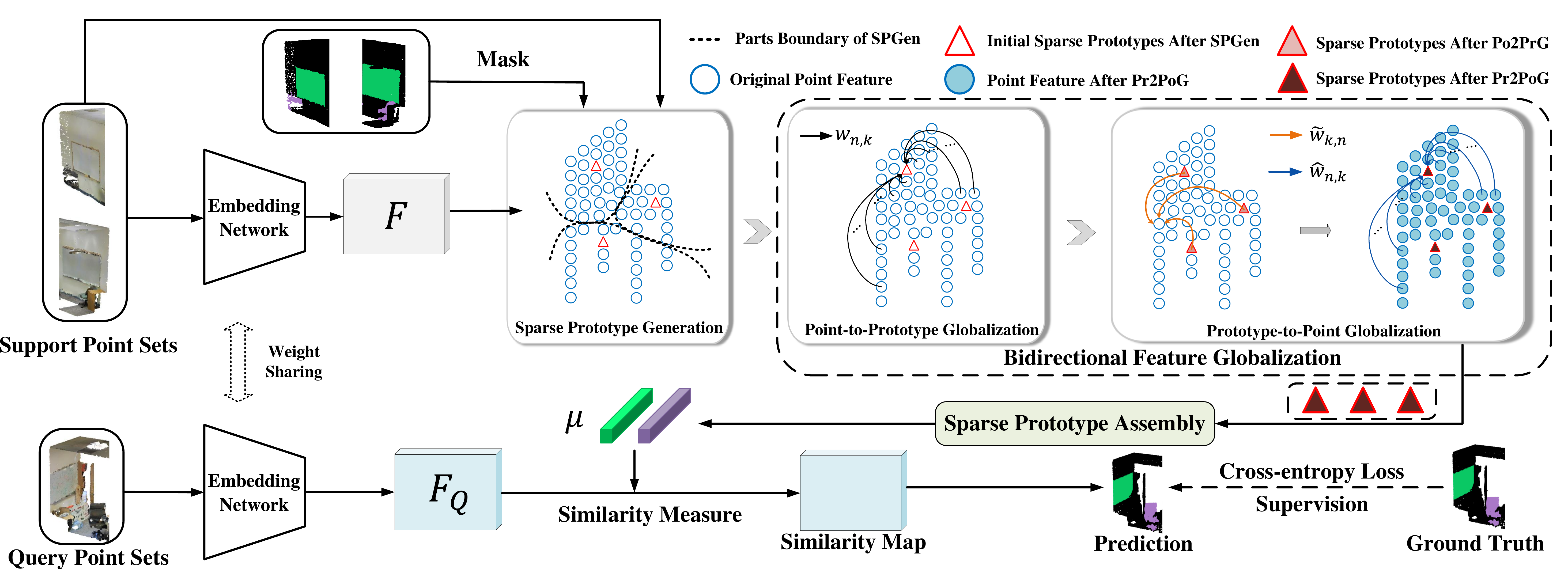}
    \end{center}
    \setlength{\abovecaptionskip}{0pt}
    \caption{Flowchart of our BFG approach. 
    The support branch is responsible for sparse prototype generation (SPGen), bidirectional feature globalization (BFG), and sparse prototype assembly (SPA).
    The query branch calculates the similarity maps between the query point features and the prototypes, and predicts the segmentation results.}
    \label{framework}
\end{figure*}

The flowchart of our BFG approach is illustrated in Fig.~\ref{framework}, which uses ProtoNet~\cite{dong2018few} as the baseline.
As a few-shot 3D segmentation network, BFG consists of two network branches: the support branch (upper) and the query branch (lower).
The two network branches use a weight shared feature embedding network to extract point features.
Let $F$ and $F_Q$ represent point features after passing through the embedding network of support branch and query branch, respectively.
In the support branch, the prototypes are first generated by sparse prototype generation (SPGen) upon the point features $F$ and the corresponding mask.
By passing the point-to-prototype globalization (Po2PrG) and prototype-to-point globalization (Pr2PoG) modules, these prototype vectors are endowed with global feature perception.
A sparse prototype assembly (SPA) module is applied to aggregate the prototypes for semantic representation.
In the query branch, similarity maps between the extracted features $F_Q$ of query point sets and the prototypes extracted by the support branch are calculated by the distance function (cosine distance or squared Euclidean distance). 
Such similarity maps are directly used to produce semantic segmentation results. In the query branch, the network is driven by the cross-entropy loss $\mathcal{L}_{CE}$, as $\mathcal{L}_{total}=\mathcal{L}_{CE}$.

In what follows, we first introduce the SPGen module and then present the feature globalization procedure with Po2PrG and Pr2PoG modules.

%-------------------------------------------------------------------------
\subsection{Sparse Prototype Generation}
Sparse prototype generation (SPGen) produces the initial representation of prototypes, Fig. ~\ref{framework}. For each class of support point sets, we first use the mask of points to get the foreground and background points. 
Inspired by ~\cite{jampani2018superpixel,hui2021superpoint}, the foreground points are partitioned into multiple groups which correspond to object parts. Each part corresponds to a prototype vector. 
Following~\cite{zhao2021few}, the sparse prototypes are initialized by two steps: object part construction and prototype extraction. 

\textit{Object Part Construction. }
Denote $N$ and $D$ as the number and the channel of point features, respectively. Given the support point feature $F\in \mathbb{R}^{N\times D}$, the coordinate $J\in \mathbb{R}^{N\times3}$, and the mask $M^c\in \mathbb{R}^{N\times{1}}$ ($c$ represents the class) of support point sets, the masked point feature $\mathcal{F}^{c}=\{f^{c}_i\}_{i=1}^{m^{c}}$ and its coordinate $\mathcal{J}^{c}=\{j_i^{c}\}_{i=1}^{m^{c}}$ ($m^{c}$ represents the number of support points belonging to the class $c$) are obtained through its corresponding mask $M^c$ and implemented by keeping points of class $c$ and culling points of other classes.

Based on the point feature $\mathcal{F}^{c}$ and coordinate $\mathcal{J}^{c}$, sampling seed points and point-to-seed assignment
~\cite{zhao2021few} are executed sequentially. 
The farthest point sampling (FPS) algorithm is employed to sample a subset of $K$ seed points which are from the same class. 
The seed points represent the centers of the parts. Let $\{s_k^c\}_{k=1}^{K}\subset \{f_i^c\}_{i=1}^{m^c}$ denote the sampled seeds. After that, we compute the point-to-seed distance and assign point features to these part centers according to the index of the closest part center of each point.

\textit{Prototype Extraction. }
We perform global average pooling within each part to extract prototypes. 
Formally, the initial sparse prototype $\mu^{c}$ of class $c$ is defined as:
\begin{equation}
\begin{aligned}
   &\mu^{c} = \Big\{\mu^c_1,\cdots ,\mu^c_K|\mu^c_k = \frac{1}{|\mathcal{I}^c_k|}\sum_{f^c_i\in \mathcal{I}^c_k}f^c_i\Big\},\\
   &\ s.t.\ \underset{\mathcal{I}^c}{\mathrm{argmin}}\sum_{k=1}^K\sum_{f^c_i\in \mathcal{I}_k^c}||f^c_i-s^c_k||_2
\end{aligned}
\end{equation}
where the masked point features $\mathcal{F}^{c}=\{f^{c}_i\}_{i=1}^{m^{c}}$ is partitioned to $K$ sets $\mathcal{I}^c=\{\mathcal{I}^c_1,\cdots,\mathcal{I}^c_K\}$ such that $f^c_i\in \mathcal{I}^c_k$ is assigned to $s_k^c$. At the same time, the coordinate of each point $j^c_i\in \mathcal{I}^c_{\mathcal{J},k}$ is also assigned to $s_k^c$. In this way, we get the coordinates of the part sets as $\mathcal{I}^c_\mathcal{J}=\{\mathcal{I}^c_{\mathcal{J},1},\cdots,\mathcal{I}^c_{\mathcal{J}, K}\}$. 
Similarly, the coordinate of each prototype is defined as:
\begin{equation}
\begin{aligned}
   &\mu^{c}_{\mathcal{J}} = \Big\{\mu^c_{\mathcal{J},1},\cdots ,\mu^c_{\mathcal{J},K}|\mu^c_{\mathcal{J},k} = \frac{1}{|\mathcal{I}^c_{\mathcal{J}, k}|}\sum_{j^c_i\in \mathcal{I}^c_{\mathcal{J},k}}j^c_i\Big\}
\end{aligned}
\end{equation}

\subsection{Bidirectional Feature Globalization}
\begin{figure*}
    \begin{center}
    \includegraphics[scale=.35]{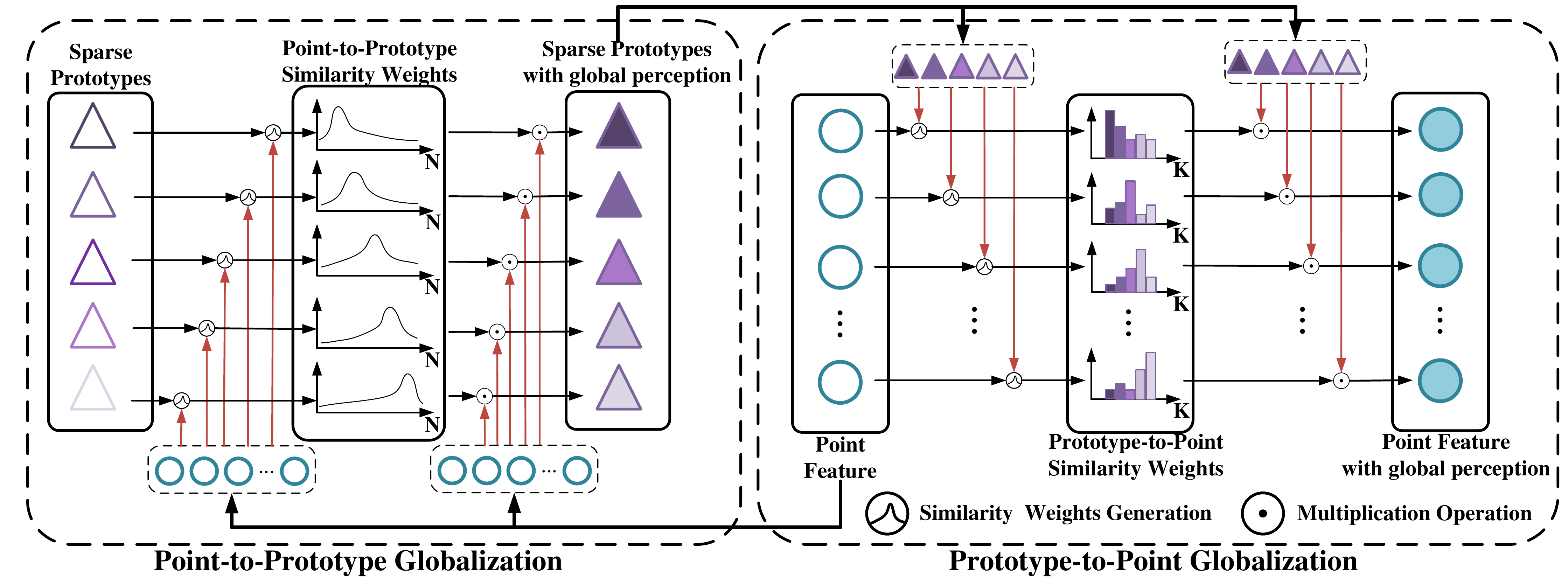}
    \end{center}
    \setlength{\abovecaptionskip}{0pt}
    \caption{Diagram of point-to-prototype globalization (Po2PrG) and prototype-to-point globalization (Pr2PoG). 
    Prototypes and point features are enhanced with global perception by generating similarity weights in a bidirectional fashion.}
    \label{BidirectionalGlobalization}
\end{figure*}

Since the initialized prototypes are extracted within object parts, the prototype semantics are limited to local point features.
To solve this issue, we propose point-to-prototype globalization (Po2PrG) to perform global representation of sparse prototypes, Fig.~\ref{BidirectionalGlobalization}(left). 
Similarly, due to the locality of point convolution, dense point features extracted by the embedding network ignore the global semantic perception between object parts.
Prototype-to-point globalization (Pr2PoG) is proposed to solve this problem, Fig.~\ref{BidirectionalGlobalization}(right).

The Po2PrG and Pr2PoG modules leverage the semantic perception of point-to-prototype and prototype-to-point similarity, respectively.
With Po2PrG and Pr2PoG, BFG embeds the global perception to both prototype vectors and point features in a bidirectional fashion. 

Before introducing Po2PrG and Pr2PoG, we define the similarity measurement for the generation of similarity weights. 
To leverage the spatial information, the coordinate $\mathcal{J}^{c}$ is introduced to the similarity measurement. Thus, the similarity between point features $\mathcal{F}^c$ and sparse prototypes $\mu^c$ can be defined as:
\begin{equation}
f(\mathcal{F}^{c},\mu^{c};\mathcal{J}^{c},\mu^c_{\mathcal{J}})
=e^{-\mathcal{D}(\mathcal{F}^{c},\mu^{c};\ \mathcal{J}^{c}, \mu^c_{\mathcal{J}})},    
\end{equation}
where $\mathcal{D}(\cdot)$ denotes the distance measurements defined on either $L_2$-Norm or inner product operation in what follows. 

(1) \textit{$L_2$-Norm} is commonly used to calculate the similarity between two feature vectors. With this operation, $\mathcal{D}(\cdot)$ is defined as:
\begin{math}
\mathcal{D}=\sqrt{d(\mathcal{F}^{c},\mu^{c})+d(\mathcal{J}^{c}, \mu^c_{\mathcal{J}})},
\end{math}
where $d(\mathcal{F}_{n}^{c},\mu^c_k)=\frac{\lambda}{max(\mathcal{F}^c_n)}\sum_{i=1}^{D}||\mathcal{F}^{c}_{n,i}-\mu^c_{k,i}||^{2}$, $d(\mathcal{J}_{n}^{c}, \mu^c_{\mathcal{J},k})=\sum_{l=1}^{3}||\mathcal{J}^{c}_{n,l}-\mu^c_{\mathcal{J},k,l}||^{2}$, and $||\cdot||$ denotes the $L_2$-Norm. $\lambda$ is the scale factor that keeps $\mathcal{F}^c_n$ feature-based distances and $\mathcal{J}^c_n$ coordinate-based distances orders of magnitude consistent and set to $0.85$ in our experiments.  

(2) \textit{Inner product operation} is formulated as:
\begin{math}
\mathcal{D}=\xi \cdot\Big[d(\mathcal{F}^{c},\mu^c)+d(\mathcal{J}^{c},\mu^c_{\mathcal{J}})\Big],
\end{math}
where $d(\mathcal{F}^c_n,\mu^c_k)={\mu^c_k}^{T}\mathcal{F}^{c}_{n}$ and $(\cdot)^T$ is the transpose operation of the matrix.
Similarly, $d(\mathcal{J}^{c}_{n},\mu^c_{\mathcal{J},k})={\mu^c_{\mathcal{J},k}}^{T}\mathcal{J}^c_n$. $\xi$ denotes the concentration factor, which is set to $0.5$ in our experiments.

\subsubsection{Point-to-Prototype Globalization.}
Point-to-Prototype globalization is carried out based upon the similarity weights from dense point features to sparse prototypes. 
Specifically, similarity measurement, weights generation, and prototype globalization are carried out in order.

\textit{Similarity Measurement.}
Given the initialized prototypes $\{\mu^c, \mu^c_{\mathcal{J}}\}$, we compute the similarity between point features and prototype vectors, as $f_1(\mathcal{F}^{c},\mu^c;\mathcal{J}^c, \mu^c_{\mathcal{J}})=e^{-\mathcal{D}_{1}}$. In experiments, $\mathcal{D}_{1}$ is defined as $L_2$-Norm. 

\textit{Weights Generation.}
In this procedure, we convert the dense point features into the weights $w_{n,k}$, which is defined on the similarity between the point features and each prototype vector, as:
\begin{equation}
\begin{aligned}
    w_{n,k}=
    \frac{f_{1;n,k}(\mathcal{F}^c,\mu^c;\mathcal{J}^c, \mu^c_{\mathcal{J}})}
    {\sum_{n=1}^{N}f_{1;n,k}(\mathcal{F}^c,\mu^c;\mathcal{J}^c, \mu^c_{\mathcal{J}})},
\end{aligned}
\end{equation}
where $f_{1;n,k}(\mathcal{F}^c,\mu^c;\mathcal{J}^c, \mu^c_{\mathcal{J}})$ is the similarity between the $n$-th point feature $\mathcal{F}_n^c$ and the $k$-th prototype vector $\mu^c_k$. Such similarity weights represent the semantic similarity between prototypes and point features. A larger weight value means higher semantic similarity.

\textit{Prototype Globalization.}
After generating the semantic similarity, prototype globalization is implemented by calculating the weighted average of the point features through the similarity weights. The new prototype $\upsilon^c_k$ is formulated as:
\begin{equation}
\begin{aligned}
   \upsilon^c_k=\sum_{n=1}^{N}w_{n,k}\mathcal{F}_n^c.
\end{aligned}
\end{equation}

With the weighted average of point features, the global perception is embedded to the sparse prototypes.

%-------------------------------------------------------------------------
\subsubsection{Prototype-to-Point Globalization.}
After Po2PrG, the sparse prototypes have acquired global perception. However, the local point features remain local dependency. 
To solve it, Pr2PoG based on similarity weights is introduced to embed global perception to local point features. This is implemented through four steps: similarity measurement, weights generation, point feature globalization, and prototype globalization.

\textit{Similarity Measurement.}
Given sparse prototypes $\upsilon^c=\{\upsilon^c_k\}_{k=1}^{K}$ which incorporate global perception, we compute the similarity between point features and global prototypes, as $f_2(\mathcal{F}^c,\upsilon^c;\mathcal{J}^c, \upsilon^c_{\mathcal{J}})=e^{-\mathcal{D}_{2}}$ where $\upsilon^c_{\mathcal{J}}=\mu^c_{\mathcal{J}}$. 
In experiments, $\mathcal{D}_{2}$ is defined as the inner product operation.

\textit{Weights Generation. }
With the similarity measurement, we correlate the global prototypes with local point features through the semantic perception $i.e.$, the similarity weights from global prototypes to dense point features.
The similarity weight $\widetilde{w}_{k,n}$ from the $k$-th prototype to the $n$-th point feature is defined as:
\begin{equation}
\begin{aligned}
   \widetilde{w}_{k,n}=
   \frac{f_{2;k,n}(\mathcal{F}^c,\upsilon^c;\mathcal{J}^c, \upsilon^c_{\mathcal{J}})}
   {\sum_{k=1}^{K}f_{2;k,n}(\mathcal{F}^c,\upsilon^c;\mathcal{J}^c, \upsilon^c_{\mathcal{J}})}.
\end{aligned}
\end{equation}

In the weights space, a prototype with higher similarity to the point feature is assigned a higher weight, and vice versa.

\textit{Point Feature Globalization. }
Based on the semantic similarity, the updated point features $\mathfrak{F}^c_n$ are obtained by the weighted average through the similarity weights, as:
\begin{equation}
\begin{aligned}
   \mathfrak{F}^c_n = \mathcal{F}_n^c +  \sum_{k=1}^{K}\widetilde{w}_{k,n}\upsilon^c_k.
\end{aligned}
\end{equation}

\textit{Prototype Globalization.}
To globalize the prototypes, we employ weights generation to convert the updated point features into new weights $\widehat{w}_{n,k}$, which is formulated as:
\begin{equation}
\begin{aligned}
   \widehat{w}_{n,k}=
   \frac{f_{2;n,k}(\mathfrak{F}^c,\upsilon^c;\mathcal{J}^c, \upsilon^c_{\mathcal{J}})}
   {\sum_{n=1}^{N}f_{2;n,k}(\mathfrak{F}^c,\upsilon^c;\mathcal{J}^c, \upsilon^c_{\mathcal{J}})}.
\end{aligned}
\end{equation}

The prototype globalization of Po2PrG is then used to compute the weighted average through the similarity weights and obtain enhanced prototypes $r^c_k$, as:
\begin{equation}
\begin{aligned}
   r^c_k=\sum_{n=1}^{N}\widehat{w}_{n,k}\mathfrak{F}^c_n.
\end{aligned}
\end{equation}

After Pr2PoG, point features are correlated with the sparse prototypes and the enhanced prototypes are obtained for the following prototype assembly.

\begin{table*}[htb]
    \centering
    \setlength{\abovecaptionskip}{0.cm}
    \setlength{\belowcaptionskip}{-0.cm}
\scriptsize
    \begin{center}
    \caption{Performance on S3DIS. `Embed. Net' denotes the embedding network (backbone). `DGCNN w/o SAN' and `DGCNN w/ SAN' denote DGCNN backbone without and with SAN, respectively.  S$^{i}$ represents the split $i$ is selected to test our BFG.} \label{tables3dis}
    \begin{tabular}{l|c|cc>{\columncolor{mygray}}c|cc>{\columncolor{mygray}}c|cc>{\columncolor{mygray}}c|cc>{\columncolor{mygray}}c}
    \hline
    \multirow{3}*{Method}&\multirow{3}*{Embed. Net}     
          & \multicolumn{6}{c|}{2-way}                                    & \multicolumn{6}{c}{3-way}                                  \\ \cline{3-14}
    ~ &~    & \multicolumn{3}{c|}{1-shot}    & \multicolumn{3}{c|}{5-shot}  & \multicolumn{3}{c|}{1-shot}  & \multicolumn{3}{c}{5-shot}  \\ \cline{3-14}
    ~ &~    & S$^{0}$ & S$^{1}$ & \bf{mean}       & S$^{0}$ & S$^{1}$ & \bf{mean}     & S$^{0}$ & S$^{1}$ & \bf{mean}     & S$^{0}$ & S$^{1}$ &\bf{mean}     \\
    \hline
    \hline
   FT~\cite{zhao2021few}      &\multirow{4}*{DGCNN w/o SAN}   & 36.34  & 38.79 & 37.57 & 56.49 & 56.99 & 56.74 & 30.05 & 32.19 & 31.12 & 46.88 & 47.57 & 47.23 \\
   ProtoNet~\cite{dong2018few} &~  & 48.39  & 49.98 & 49.19 & 57.34 & 63.22 & 60.28 & 40.81 & 45.07 & 42.94 & 49.05 & 53.42 & 51.24 \\
   MPTI~\cite{zhao2021few}     &~  & 52.27  & 51.48 & 51.88 & 58.93 & 60.56 & 59.75 & 44.27 & 46.92 & 45.60 & 51.74 & 48.57 & 50.16 \\
  \bf{BFG(ours)}&~  & 52.50 &53.26  & \bf{52.88} &59.26 & 62.82  &\bf{61.04} &43.80 & 47.76 &\bf{45.78} & 49.80 & 55.10 &\bf{52.45} \\
    \hline
    \hline
   ProtoNet~\cite{dong2018few} &\multirow{3}*{DGCNN w/ SAN} & 50.98  & 51.90 & 51.44 & 61.02 & 65.25 & 63.14 & 42.16 & 46.76 & 44.46 & 52.20 & 56.20 & 54.20 \\
   MPTI~\cite{zhao2021few}   &~  & 53.77  & 55.94 & 54.86 & 61.67 & 67.02 & 64.35 & 45.18 & 49.27 & 47.23 & 54.92 & 56.79 & 55.86 \\
  \bf{BFG(ours)}  &~  & 55.60  & 55.98 & \bf{55.79} & 63.71  &66.62 & \bf{65.17} &46.18 &48.36 & \bf{47.27} &55.05 & 57.80  &\bf{56.43} \\
   \hline
\end{tabular}
\end{center}
\end{table*}
\begin{table*}[htb]
   \setlength{\abovecaptionskip}{0.cm}
    \setlength{\belowcaptionskip}{-0.cm}
   \scriptsize
   \begin{center}
   \caption{Performance on ScanNet. `Embed. Net' denotes the embedding network (backbone). `DGCNN w/o SAN' and `DGCNN w/ SAN' denote DGCNN backbone without and with SAN, respectively. S$^{i}$ represents the split $i$ is selected to test our BFG.}
   \label{tablescannet}
   \begin{tabular}{l|c|cc>{\columncolor{mygray}}c|cc>{\columncolor{mygray}}c|cc>{\columncolor{mygray}}c|cc>{\columncolor{mygray}}c}
   \hline
   \multirow{3}*{Method} &\multirow{3}*{Embed. Net}   
         & \multicolumn{6}{c|}{2-way}                                    & \multicolumn{6}{c}{3-way}                                  \\ \cline{3-14}
   ~  &~   & \multicolumn{3}{c|}{1-shot}    & \multicolumn{3}{c|}{5-shot}  & \multicolumn{3}{c|}{1-shot}  & \multicolumn{3}{c}{5-shot}  \\ \cline{3-14}
   ~  &~     & S$^{0}$ & S$^{1}$ & \bf{mean}       & S$^{0}$ & S$^{1}$ &\bf{mean}     & S$^{0}$ & S$^{1}$ & \bf{mean}     & S$^{0}$ & S$^{1}$ &\bf{mean}     \\
   \hline
   \hline
   FT~\cite{zhao2021few}      &\multirow{4}*{DGCNN w/o SAN}    & 31.55  & 28.94 & 30.25 & 42.71 & 37.24 & 39.98 & 23.99 & 19.10 & 21.55 & 34.93 & 28.10 & 31.52  \\
   ProtoNet~\cite{dong2018few}  &~  & 33.92  & 30.95 & 32.44 & 45.34 & 42.01 & 43.68 & 28.47 & 26.13 & 27.30 & 37.36 & 34.98 & 36.17  \\
   MPTI~\cite{zhao2021few}     &~    & 39.27  & 36.14 & 37.71 &46.90 & 43.59 & \bf{45.25} & 29.96 & 27.26 & 28.61 & 38.14 & 34.36 & 36.25  \\
  \bf{BFG(ours)}&~  &38.63   &36.82 & \bf{37.73}  & 45.67 &42.36 &44.02 &30.57 &29.02 &\bf{29.80} &38.64 &34.75 &\bf{36.70}\\
   \hline
   \hline
   ProtoNet~\cite{dong2018few}&\multirow{3}*{DGCNN w/ SAN} & 37.99  & 34.67 & 36.33 & 52.18 & 46.89 & 49.54 & 32.08 & 28.96 & 30.52 & 44.49 & 39.45 & 41.97  \\
   MPTI~\cite{zhao2021few}  &~ & 42.55  & 40.83 & \bf{41.69} & 54.00 & 50.32 & \bf{52.16} & 35.23 & 30.72 & 32.98 & 46.74 & 40.80 & 43.77  \\
\bf{BFG(ours)}&~   &42.15 &40.52 & 41.34 &51.23 &49.39 & 50.31 &34.12 &31.98 &\bf{33.05} &46.25 &41.38 & \bf{43.82} \\
   \hline
\end{tabular}
\end{center}
\end{table*}
%-------------------------------------------------------------------------
\subsection{Sparse Prototype Assembly}
After prototype globalization, we obtain the enhanced prototypes with global semantic perception, as $r^c=\{r^c_k\}_{k=1}^{K}$.
When performing semantic segmentation in the following metric learning procedure, it requires to generate the similarity maps between query features and prototype vectors.
To perform the measurement, the sparse prototypes from each class require to be fused at first. 
We first apply MLP to the prototypes, as:
\begin{math}
    \widehat{r}^{c}=\mathcal{M}\big\{r^c\big\}\in \mathbb{R}^{D\times K}
\end{math}.
We then calculate the mean prototype $z^c$ of each class, as
\begin{equation}
\begin{aligned}
   z^{c}=\sum_{k=1}^{K}\alpha_{k}\circ \widehat{r}^{c}_k,
\end{aligned}
\end{equation}
where \begin{math}\alpha_{k}=\frac{e^{\widehat{r}^{c}_k}}{\sum_{i=1}^{K}e^{\widehat{r}^{c}_i}}\in \mathbb{R}^{D\times 1}\end{math} is the normalized weight vector of the prototypes for class $c$, and $\circ$ is the Hadamard product. The prototypes of all classes are $z =\{z^{c}\}_{c=1}^{C}$. 

%-------------------------------------------------------------------------
\section{Experiments}
The proposed BFG is evaluated on two point cloud segmentation benchmarks, including the Stanford Large-Scale 3D Indoor Spaces (S3DIS)~\cite{armeni20163d} and ScanNet~\cite{dai2017scannet}. 

%-------------------------------------------------------------------------
\subsection{Datasets}
\textbf{S3DIS. }
S3DIS is a dataset which collects 3D RGB point clouds from 272 rooms in six indoor environments. Each point is annotated with one of the semantic labels from 13 classes (12 semantic classes plus the clutter). 
\textbf{ScanNet.}
The ScanNet dataset contains a total of 1513 scanned scenes. All the points except the unannotated space are annotated by 20 semantic classes. 

\textbf{Data pre-processing.}
Following ~\cite{zhao2021few}, we divide the semantic classes of each dataset into two non-overlapping combinations $S^0$ 
(S3DIS: \textit{beam, board, bookcase, ceiling, chair, column.} ScanNet: \textit{bathtub, bed, bookshelf, cabinet, chair, counter, curtain, desk, door, floor}) 
and $S^1$ 
(S3DIS: \textit{door, floor, sofa, table, wall, window.} ScanNet: \textit{other furniture, picture, refrigerator, show curtain, sink, sofa, table, toilet, wall, window}) 
according to the alphabetical order. 
To facilitate the point cloud be fed to the network, we process the datasets following ~\cite{qi2017pointnet,zhao2021few}: splitting the rooms into $1m\times 1m$ blocks and sampling 2048 points from the block each time. 
After that, the S3DIS and ScanNet datasets are split into 7547 and 36350 blocks, respectively. 
Since the area of each block is small, the sampled points can only contain one instance object or a local area of an instance object. 
For each dataset, the cross-validation is performed to our method, which is implemented by selecting one split $S^{i}$ as the train class set and regarding the other split $S^{1-i}$ as the test class set. 
If $S^i$ ($S^0$ or $S^1$) is used as the test class set, the blocks containing the $S^i$ class are selected as the test set, and the blocks containing the $S^{1-i}$ class is selected as the training set. 
Following ~\cite{zhao2021few}, the sampling process of each episode in the training process is as follows: 
(1) Randomly selecting a combination of $N$ classes from the train class set $S^{i}$ to set up $N$-way as the foreground, and the remaining classes are regarded as the background. 
(2) Randomly sampling $K$ ($K$-shot) support sets and query sets based on the selected $N$-way class combination in the training set.   
When testing, we take the same operation to preprocess the data of the test class set.
\begin{figure*}[t]
    \begin{center}
    \includegraphics[width=1.0\linewidth]{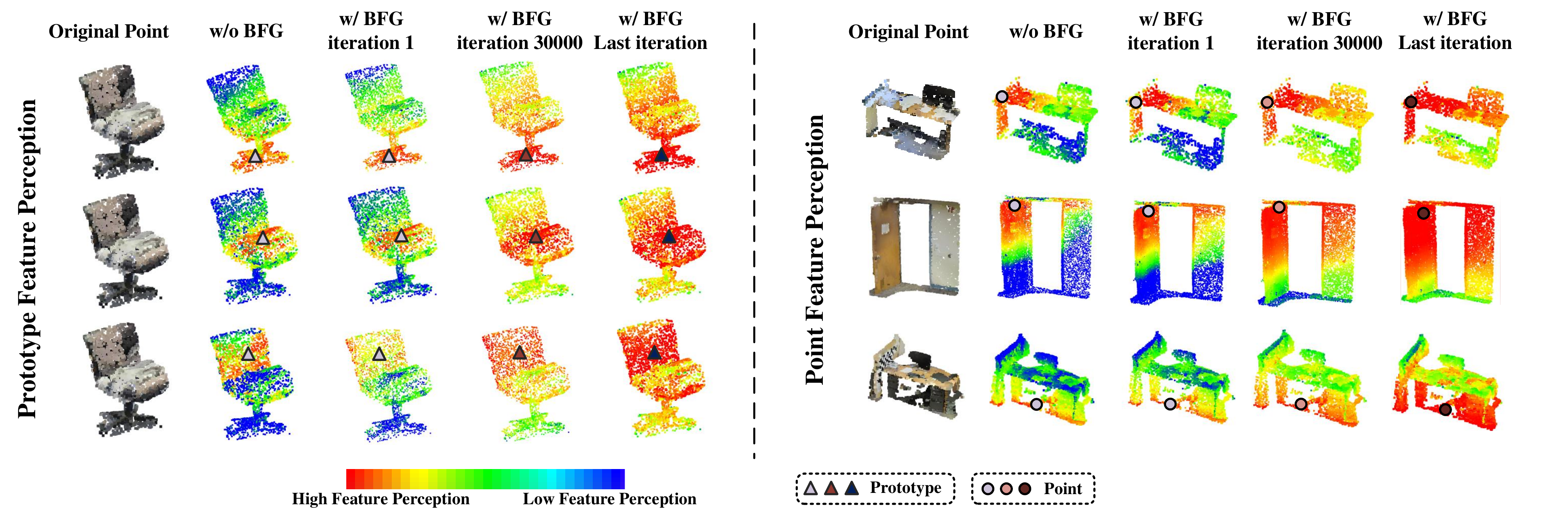}
    \end{center}
    \setlength{\abovecaptionskip}{0pt}
      \caption{Feature perception of prototypes and point features during training. 
    Both prototype vectors and local point features with BFG are gradually embedded with global perception, which extends the receptive fields to full instance extent. (Best viewed in color)}
\label{VisualGlobalization}
\end{figure*}

%-------------------------------------------------------------------------
\subsection{Experimental Settings}
\textbf{Evaluation Protocols. }
Mean intersection over union (mIoU), which is widely used in point cloud segmentation, is selected as the metric for performance evaluation.

\textbf{Implementation Details. }
DGCNN ~\cite{wang2019dynamic} is proposed as a basic point cloud classification and segmentation network that is widely used for many point cloud processing tasks. SAN~\cite{zhao2021few} is introduced to explore correlations of semantic context within the point set. Based on this, DGCNN (without or with SAN) is selected as the backbone. 
Our approach utilizes ProtoNet~\cite{zhao2021few} as the baseline. 
Following~\cite{zhao2021few}, Gaussian jittering operation and random rotation operation around z-axis data augmentation strategies are used in the training process. 
Our approach runs on a single NVIDIA TITAN RTX GPU with the batch size set to 1 and the number of training iterations set to 80000. 
The Adam optimizer is used with an initial learning rate of 0.0001 for the embedding network. 
Furthermore, an initial learning rate for the remaining parts is set to 0.001. 
Since S3DIS and ScanNet belong to the same type of indoor scene point cloud data, the number of sparse prototypes is set to 5 on both S3DIS and ScanNet through extensive experiments.

%-------------------------------------------------------------------------
\subsection{Performance}
\textbf{S3DIS. }
In Table ~\ref{tables3dis}, our BFG is compared with the state-of-the-art methods. 
DGCNN without and with SAN are selected as the backbone, BFG outperforms state-of-the-art methods in all experimental settings.
With the 2-way 1-shot setting and DGCNN (with SAN) backbone, our BFG achieves $\bf{4.35\%}$ ($55.79\%$ vs. $51.44\%$) performance improvement over the baseline ProtoNet. 
Note that for the $S^0$ setting in 2-way 1-shot experiments, BFG improves the baseline ProtoNet by $\bf{4.62\%}$, which outperforms the state-of-the-art method MPTI by $\bf{1.83\%}$.
Compared with ProtoNet, it is concluded that the extraction of sparse prototypes does help to improve the segmentation performance.
Compared with MPTI, we can see that too many prototypes will not have higher performance. 
On the contrary, a moderate number of prototypes with the optimal global representation after BFG greatly improve the performance.

\textbf{ScanNet. }
Table ~\ref{tablescannet} displays the segmentation performance on ScanNet. 
BFG again outperforms state-of-the-art methods in most experimental settings. For the 2-way 1-shot setting and DGCNN (without SAN) backbone, BFG improves the baseline ProtoNet by $\bf{5.29\%}$ ($37.73\%$ vs. $32.44\%$), which is a large margin for the challenging few-shot 3D segmentation problem. 
Compared to S3DIS, there are more classes and point sets in ScanNet, which facilities learning richer representation related to various instances. Thereby, the improvement on 2-way 1-shot setting of ScanNet is larger than that on S3DIS. 
With the DGCNN (with SAN) backbone, our BFG is better than or on par with state-of-the-arts.

%-------------------------------------------------------------------------
\subsection{Visualization Analysis}
\textbf{Feature Perception. }
As Fig.~\ref{VisualGlobalization} shows, we visualize the feature perception of different prototypes and point features during training, respectively. 
In the left part, the feature perception of different prototypes in the same class without and with BFG is clearly displayed.
Prototype feature perception in Fig.~\ref{VisualGlobalization} is reflected by the similarity between the point features and each prototype. 
In the right part, the feature perception of point features without and with BFG is given.
Point feature perception in Fig.~\ref{VisualGlobalization} is reflected by the point feature maps before and after adding BFG. 
One can see that with BFG, both the prototypes and the point features gradually approach global perception. 
From the visualizations, the feature perception of both the prototypes and the point features after our BFG can be extended to full instance extent.
These clearly demonstrate the superiority of our BFG in feature representation over previous prototype methods.

\begin{figure*}[t]
    \begin{center}
    \includegraphics[width=1.0\linewidth]{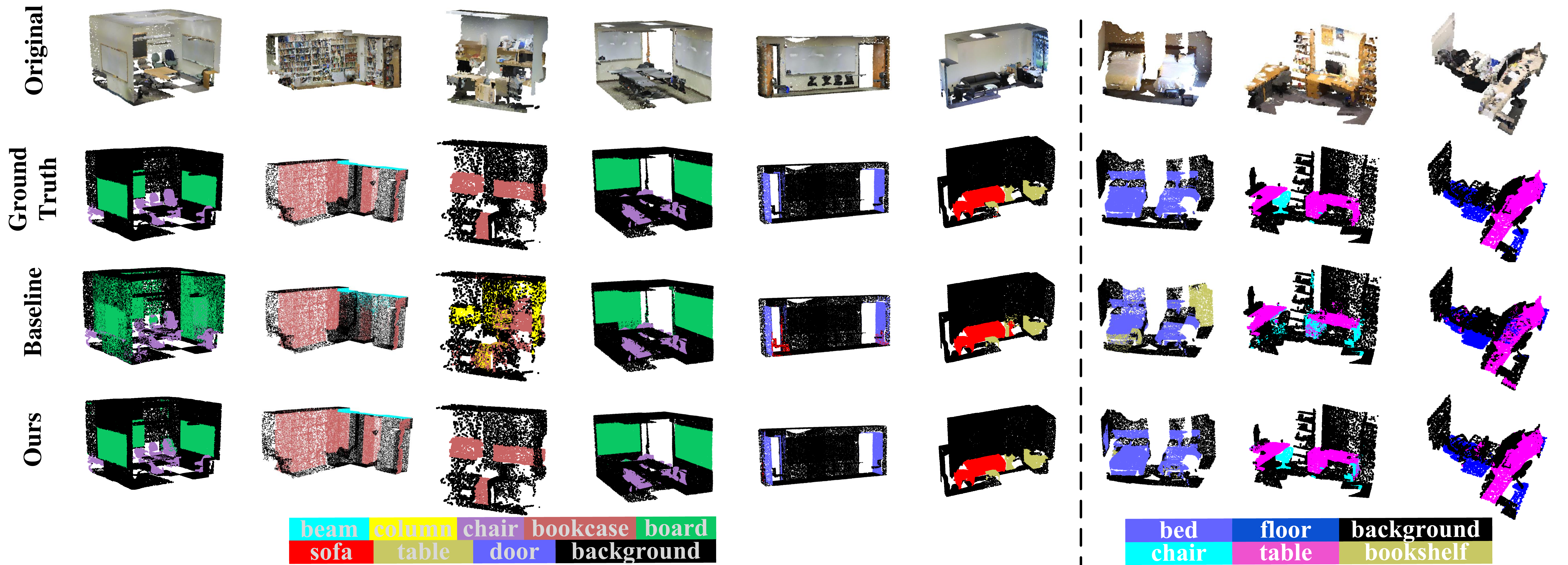}
    \end{center}
    \setlength{\abovecaptionskip}{0cm}
    \setlength{\belowcaptionskip}{-0.cm}
      \caption{Visualizations of segmentation results on S3DIS and ScanNet. Left: S3DIS. Right: ScanNet. }
\label{s3dis}
\end{figure*}

\textbf{Segmentation Results. } 
As Fig.~\ref{s3dis} shows, we compare the segmentation performance by the baseline method and our BFG on S3DIS and ScanNet datasets.
The segmentation results show that by introducing bidirectional feature globalization, BFG has achieved good segmentation performance on the instances with scale variation and deformation. 
In addition, the optimal sparse prototypes also greatly alleviate the false segmentation between classes.

%-------------------------------------------------------------------------
\subsection{Ablation Study}
We conducted ablation studies with 
2-way 1-shot setting on the S3DIS dataset to verify the effectiveness of BFG. 
ProtoNet~\cite{dong2018few} is selected as the baseline. 
\begin{table}[htb]
\setlength{\abovecaptionskip}{0.cm}
\setlength{\belowcaptionskip}{-0.cm}
  \begin{center}
  \caption{Ablation study of modules in our BFG approach. The first row is the performance of the baseline ProtoNet. `SPGen' denotes the sparse prototype generation module with Sparse Prototype Assembly, `Po2PrG' denotes Point-to-Prototype Globalization, and `Pr2PoG' denotes Prototype-to-Point Globalization.}\label{tableablation}
   \begin{tabular}{ccc|ccc}
   \hline
    +SPGen &+Po2PrG & +Pr2PoG &Mean &$\Delta$ &$\Sigma\Delta$ \\
   \hline
                  &            &            &51.44      &      &      \\
      \checkmark  &            &            &53.34      &1.90  &1.90  \\
      \checkmark  & \checkmark &            &54.39      &1.05  &2.95  \\
      \checkmark  & \checkmark & \checkmark &\bf{55.79} &1.40  &4.35  \\
   \hline
\end{tabular}
\end{center}
\end{table}

\begin{figure}[t]
    \setlength{\abovecaptionskip}{0.cm}
    \setlength{\belowcaptionskip}{-0.cm}
    \begin{center}
    \includegraphics[width=1.0\linewidth]{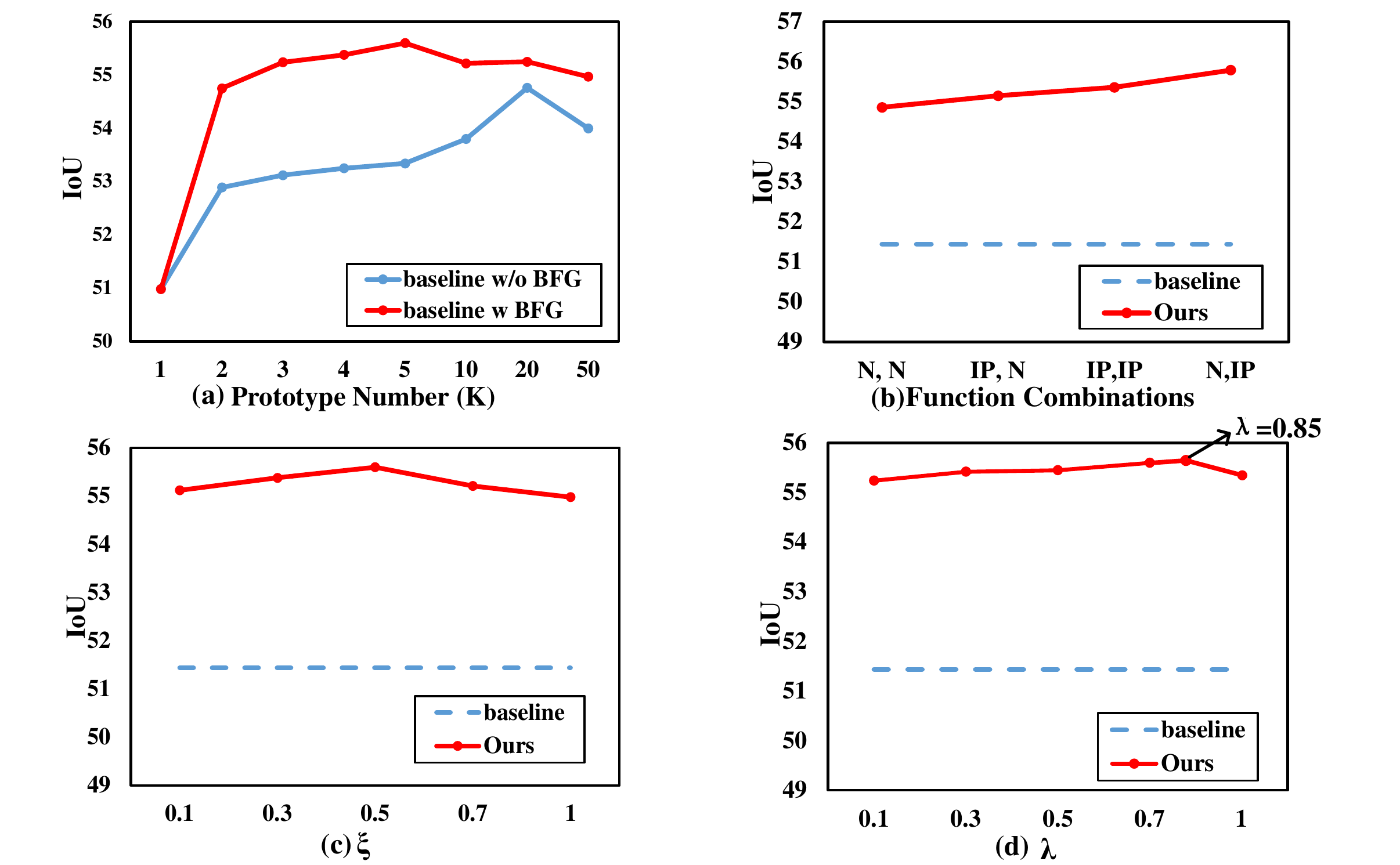}
    \end{center}
    \caption{Ablation study of the hyper-parameters and modules. (a) Prototype number $K$. (b) Distance measurement function  combinations (`N' represents $L_2$-Norm and `IP' represents inner product operation). (c) Hyper-parameter $\xi$. (d) Hyper-parameter $\lambda$.}
\label{numk}
\end{figure}

\textbf{SPGen. }
As shown in Table ~\ref{tableablation}, with sparse prototype generation, BFG improves the segmentation performance by 
$\bf{1.90}\%$ ($53.34\%$ vs. $51.44\%$) on mean IoU of the $S^0$ and $S^1$ split, which validates that multiple prototypes improve segmentation performance more significantly.
To maximizing the performance gain, a proper number of prototypes should be used.
As shown in Fig.~\ref{numk}(a), 20 prototypes can reach the best performance.

\textbf{Po2PrG. }
In Table ~\ref{tableablation}, Po2PrG further improves the performance by $\bf{1.05}\%$ ($54.39\%$ vs. $53.34\%$), which validates the prototypes acquire global perception after Po2PrG embeds local point features to the prototypes.

\textbf{Pr2PoG. }
As shown in Table ~\ref{tableablation}, by using Pr2PoG, our BFG improves the segmentation performance by $\bf{1.40}\%$ ($55.79\%$ vs. $54.39\%$), validating that the local point features obtain the global perception through embedding the global prototypes to the local point features.
This clearly demonstrates the superiority of our BFG over other methods on global feature representation.

\textbf{Number of Sparse Prototypes. }
Since S3DIS and ScanNet are divided into the blocks of 1m$\times$1m, the input of the network is 2048 points sampled from a small block. Therefore, for the input of the network, the point set only contains a single object or a part of a single object, and there will not be multiple objects. Thus, the theory that each of the prototypes represents a part of an object is sound enough. From the red line in Fig.~\ref{numk}(a), we validate that after adding BFG, setting the number of prototypes to $K=5$ can achieve the best performance, where the number of prototypes is within an acceptable range. 
Moreover, the value of $K$ shows an upward trend in segmentation performance within a certain interval $[1,5]$, and reaches a peak when $K=5$. 
As the $K$ value gradually increases, the segmentation performance decreases. 
Adding BFG makes the peak of the performance curve come early and greatly reduces the number of multiple prototypes, which surpasses the performance of MPTI with 100 prototypes while using just $K=5$ prototypes. 
This once again verifies that BFG can extract the optimal representation of prototypes.

\textbf{Distance Measurement. }
In Fig.~\ref{numk}(b), we compare the combination of $L_2$-Norm and inner product operation for distance measurement. The results from the combinations show that the distance measurement defined by ($L_2$-Norm, Inner Product) is preferable. 
As Fig.~\ref{numk}(c) and (d) show, we illustrate the effects of hyper-parameter $\xi$ in inner product operation and $\lambda$ in $L_2$-Norm.  For $\lambda$, the best performance of our BFG occurs at $\lambda=0.85$. Furthermore, the best performance of the parameter $\xi$ which is a numerical adjustment of the distance occurs at 0.5. 
Obviously, the segmentation performance of our BFG is insensitive to them.

%-------------------------------------------------------------------------
\section{Conclusion}
We propose bidirectional feature globalization (BFG), which improves the representation ability of prototypes by incorporating global feature perception in a bidirectional fashion.
BFG realizes the optimal representation of part-wised semantics through globalizing sparse prototypes and dense point features.
BFG implements prototype combination towards fusing part-wised object semantics. 
Extensive experiments on commonly used 3D segmentation datasets demonstrate the effectiveness of BFG, in striking contrast with other state-of-the-art methods. 
As a simple-yet-effective approach, BFG provides a fresh insight to the challenging few-shot segmentation task of point clouds.

{\small
\bibliographystyle{ieee_fullname}
\bibliography{egbib}
}

\end{document}